%% file: main.tex
\documentclass[10pt,twocolumn,letterpaper]{article}

\usepackage[pagenumbers]{cvpr}   

\usepackage{pifont}
\usepackage{xcolor}
\usepackage{fontawesome5}

\definecolor{linkblue}{HTML}{1677C8}
\definecolor{cvprblue}{rgb}{0.21,0.49,0.74}

\input{preamble}

\usepackage[
    pagebackref,
    breaklinks,
    colorlinks,
    allcolors=cvprblue
]{hyperref}

\hypersetup{
    urlcolor=linkblue
}

\def\confName{CVPR}
\def\confYear{2027}

\title{AcrossPlace: Do Vision--Language Models Recognize Places Beyond Their Source Views?\\
A Comprehensive Benchmark
}

\author{
Zhenyu Li$^{*}$, Shuhan Dong,
Ruotong Xu, Qingyu Zhu\\
Shandong Academy of Sciences (SAC)
\\[6pt]
\small
\href{https://github.com/CV4RA/AcrossPlace}{
    \textcolor{linkblue}{\faGithub\ GitHub}
}
\quad | \quad
\href{https://acrossplace.ornatesparrow2.chatgpt.site/}{
    \textcolor{linkblue}{\faGlobe\ Project Website}
}
}

\begin{document}
\makeatletter
\twocolumn[{%
\renewcommand\twocolumn[1][]{#1}%
\maketitle
\vspace{-0.5em}
\input{sec/title_figure}
\vspace{1em}
}]
\makeatother
\input{sec/0_abstract}
\input{sec/1_intro}
\input{sec/2_related}
\input{sec/3_benchmark}
\input{sec/4_setup}
\input{sec/5_results}
\input{sec/6_discussion}
\FloatBarrier\clearpage
   \bibliographystyle{ieeenat_fullname}
   \bibliography{main}
\end{document}

%% file: preamble.tex
\usepackage{graphicx,booktabs,amsmath,amssymb,tabularx,microtype,placeins}

\usepackage{colortbl}
\definecolor{rankBest}{HTML}{FFF0BF}
\definecolor{rankSecond}{HTML}{E2EFF8}
\definecolor{rankThird}{HTML}{ECE7F5}
\newcommand{\bestresult}[1]{\cellcolor{rankBest}{\bfseries\boldmath #1}}
\newcommand{\secondresult}[1]{\cellcolor{rankSecond}#1}
\newcommand{\thirdresult}[1]{\cellcolor{rankThird}#1}

%% file: sec/title_figure.tex
\begin{minipage}{\textwidth}
\centering
\includegraphics[width=0.98\linewidth]{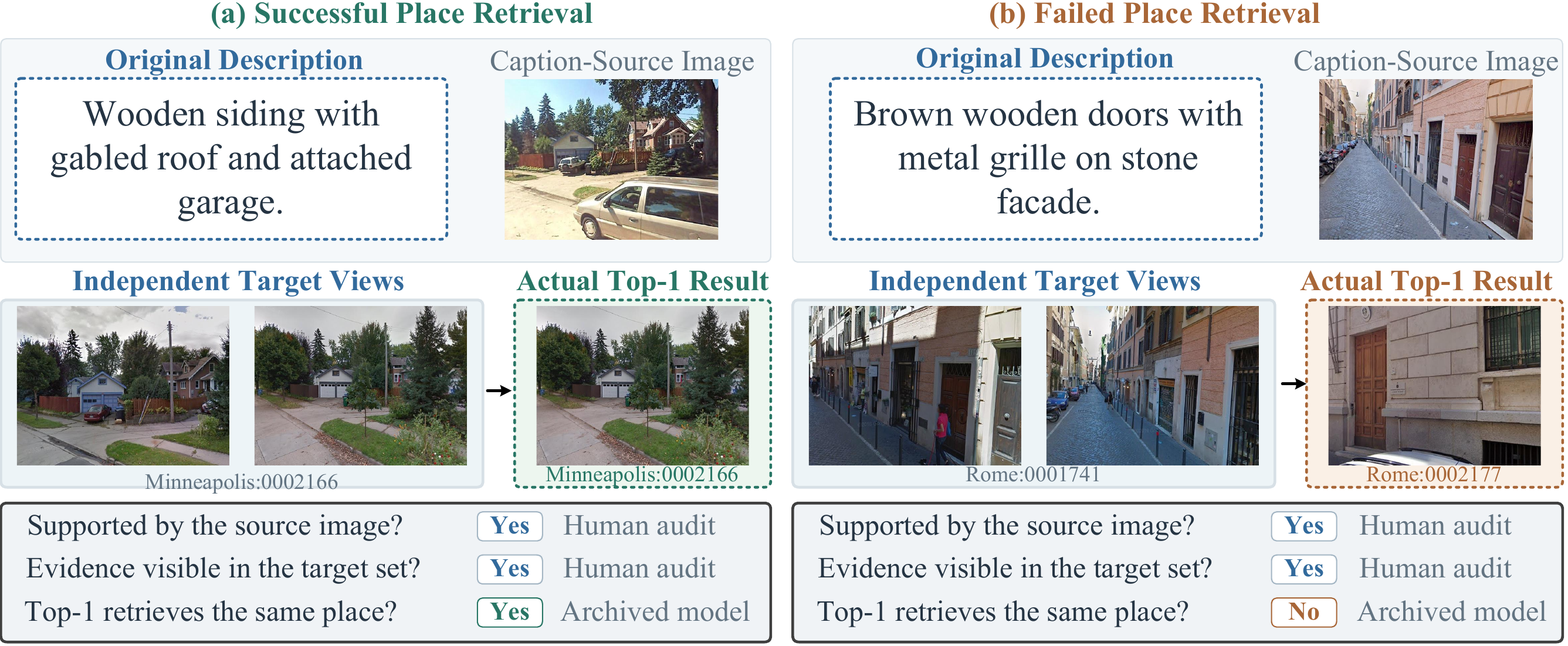}
\captionsetup{type=figure}\caption{Does recognizing visual evidence suffice to recognize a place? Verbatim captions and unmodified GSV-Cities images illustrate a successful (left) and failed (right) T2I query. Top-1 predictions are archived coarse results from place-supervised SigLIP2. Source support and target-set visibility are single-reviewer human labels; both are positive in these cases. Examples are illustrative, not frequency estimates.}\label{fig1}
\end{minipage}

%% file: sec/0_abstract.tex
\begin{abstract}
Language-guided place recognition requires recovering geographic identity across visual observations beyond the image from which a description originates. However, aggregate retrieval scores often entangle observation transfer, gallery competition, and linguistic evidence, obscuring the underlying recognition capability. We introduce \textbf{AcrossPlace}, a provenance-aware benchmark establishing a systematic evaluation standard for bidirectional language--vision place recognition. Built upon GSV-Cities, AcrossPlace comprises \textbf{56,726 places across 23 cities}, \textbf{156,800 automatically generated, rule-filtered descriptions}, and \textbf{113,404 target images}. Each description retains its source-image provenance, while retrieval operates on disjoint target observations. Text-to-image (T2I) and image-to-text (I2T) retrieval share a unified place-identity relevance criterion supporting multiple valid matches. Two geographic protocols evaluate unseen places within represented cities (P0) and generalization to held-out cities (P1). Paired observation comparisons, gallery-controlled evaluations, fixed-query scaling, and blinded annotation auditing systematically characterize retrieval performance. Across three training seeds, place-supervised SigLIP2 T2I Recall@1 decreases from \textbf{19.69\%} on source images to \textbf{12.31\%} on independent targets under P0. For fixed P1 queries, expanding the gallery from 500 to 8,425 places reduces Recall@1 from \textbf{14.67\%} to \textbf{2.87\%}. A 240-description audit further reveals discrepancies between automated acceptance and factual evidence. These findings establish AcrossPlace as a reproducible evaluation standard for assessing transferable place identity across visual observations and geographic environments.
\end{abstract}

%% file: sec/1_intro.tex
\section{Introduction}

Geographic places retain their identities across visual observations, even when their appearances change \cite{li2026a2gc, tzachor2025effovpr}. However, a language description derived from one image may accurately characterize that observation without identifying another view of the same place \cite{li2025place, shang2025bridging, shang2025mambaplace}. This distinction exposes a fundamental challenge in language--vision place recognition: \emph{matching a description to its source image does not necessarily imply recognition of the underlying place} (Figure~\ref{fig1}). Reliable place recognition therefore requires representations that transfer across observations while remaining discriminative among geographically distinct but visually similar locations.

Recent advances in multi-view text--vision matching \cite{shang2025text4vpr}, bidirectional multimodal representations \cite{xia2024uniloc}, and large-scale language supervision \cite{idan2026lavpr} have substantially expanded the scope of language-based place retrieval \cite{shang2025fourierplace}. These developments motivate a more fundamental question: \textbf{Do language--vision models capture transferable place identity beyond the observations from which descriptions originate?} Addressing this question requires disentangling three factors that jointly determine retrieval performance: \emph{observation transfer}, which reflects the consistency of place recognition across independent views; \emph{gallery competition}, which governs discrimination among candidate locations; and \emph{language-evidence reliability}, which determines whether descriptions provide visually grounded and geographically distinctive information. A rigorous benchmark must characterize these factors to establish what retrieval performance actually measures.

To this end, we introduce \textbf{AcrossPlace}, a comprehensive benchmark for provenance-aware, bidirectional language--vision place recognition. Built upon GSV-Cities \cite{alibey2022gsvcities}, AcrossPlace encompasses \textbf{56,726 places across 23 cities}, \textbf{156,800 automatically generated, rule-filtered descriptions}, and \textbf{113,404 target images}. Its central design principle is \emph{source--target independence}: each description preserves its generating-image provenance, while retrieval operates on disjoint target observations. AcrossPlace supports both text-to-image (T2I) and image-to-text (I2T) retrieval under a unified place-identity relevance criterion, allowing multiple valid matches without requiring exact image--caption correspondence. This formulation shifts evaluation from image-specific semantic agreement toward recognition of geographic identity across distinct observations.

AcrossPlace establishes two geographic evaluation protocols: \textbf{P0}, which assesses unseen places within represented cities, and \textbf{P1}, which evaluates generalization to held-out cities. Beyond these principal protocols, three controlled diagnostics systematically characterize retrieval behavior. First, paired source--target comparisons measure observation transfer under fixed descriptions, place identities, and gallery cardinality. Second, gallery-matched comparisons and fixed-query scaling examine the influence of geographic composition and candidate competition. Third, a blinded annotation audit independently assesses source-image support, target-view visibility, and place specificity. Together, these protocols establish a structured evaluation framework connecting recognition performance with its underlying visual, geographic, and linguistic conditions.

Our experiments reveal three key findings. First, replacing caption-source images with independent target observations reduces supervised SigLIP2 T2I Recall@1 from \textbf{19.69\% to 12.31\%} on P0, revealing a substantial observation-transfer gap. Second, enlarging the P1 gallery from 500 to 8,425 places reduces Recall@1 from \textbf{14.67\% to 2.87\%} under fixed queries and model parameters, demonstrating the influence of candidate competition on geographic evaluation. Third, a blinded audit of 240 descriptions identifies 166 with full source support and 148 with sufficient target evidence, revealing discrepancies between automated screening and independently assessed linguistic evidence. Collectively, these findings demonstrate that observation consistency, gallery composition, and description reliability are essential dimensions for interpreting language--vision place recognition performance.

\noindent\textbf{Our contributions are threefold:}
\begin{itemize}
    \item We establish AcrossPlace, a large-scale language--vision resource that explicitly separates description sources from retrieval targets and defines unified place-identity relevance for bidirectional cross-modal retrieval.

    \item We introduce fixed within-city and held-out-city protocols, complemented by paired observation comparisons, gallery-controlled evaluations, and fixed-query scaling to characterize place recognition across visual observations and geographic environments.

    \item Through pretrained and place-supervised baselines, controlled retrieval experiments, and independent language-evidence auditing, we identify three critical factors governing benchmark outcomes: observation transfer, candidate competition, and description reliability.
\end{itemize}

%% file: sec/2_related.tex
\section{Related Work}

\subsection{Visual Place Recognition}

Visual place recognition (VPR) \cite{li2024feature, li2024toward} identifies geographic places across observations affected by viewpoint and appearance changes and perceptual aliasing. GSV-Cities provides large-scale place-level supervision for learning discriminative visual representations \cite{alibey2022gsvcities}. Recent approaches improve recognition through cross-image correlations, multi-observation aggregation, and semantic supervision. CricaVPR models inter-image correlations to capture consistent place features \cite{lu2024cricavpr}, while HOPS aggregates reference observations into compact place signatures \cite{malone2025hops}. SemVPR transfers semantic knowledge from vision--language models to visual place representations \cite{zhang2025semvpr}. These advances motivate evaluating whether place identity remains recoverable when a linguistic description and its retrieval target arise from distinct observations.

\begin{table*}[tp]
\centering
\small
\setlength{\tabcolsep}{4.5pt}
\renewcommand{\arraystretch}{1.08}
\caption{Comparison of evaluation protocols across language--vision place recognition benchmarks. A dash denotes information not established by the cited source, rather than the absence of a capability. AcrossPlace integrates provenance-aware bidirectional retrieval, paired source--target observation controls, geographic controls, and text auditing.}
\label{tab1}
\begin{tabular}{@{}llccccccl@{}}
\toprule
Benchmark & Scale (unit) & T2I & I2T & Prov. & Paired Obs. & Geo. Ctrl. & Text Audit & Eval \\
\midrule
Street360Loc~\cite{shang2025text4vpr} & 7k locations & \textcolor{green}{$\checkmark$} & \textcolor{red}{$\times$} & \textcolor{red}{$\times$} & \textcolor{red}{$\times$} & \textcolor{green}{$\checkmark$} & \textcolor{green}{$\checkmark$} & Direct \\
UniLoc~\cite{xia2024uniloc} & 29,438 triplets & \textcolor{green}{$\checkmark$} & \textcolor{green}{$\checkmark$} & \textcolor{red}{$\times$} & \textcolor{red}{$\times$} & \textcolor{green}{$\checkmark$} & \textcolor{red}{$\times$} & Direct \\
LaVPR~\cite{idan2026lavpr} & 651,865 pairs & \textcolor{green}{$\checkmark$} & \textcolor{red}{$\times$} & \textcolor{green}{$\checkmark$} & \textcolor{red}{$\times$} & \textcolor{green}{$\checkmark$} & \textcolor{green}{$\checkmark$} & Direct \\
DlgQuest-Cities~\cite{song2026dialoguevpr} & 106,880 images & \textcolor{green}{$\checkmark$} & \textcolor{red}{$\times$} & \textcolor{green}{$\checkmark$} & \textcolor{red}{$\times$} & -- & \textcolor{green}{$\checkmark$} & Direct \\
AcrossPlace & 56,726 places & \textcolor{green}{$\checkmark$} & \textcolor{green}{$\checkmark$} & \textcolor{green}{$\checkmark$} & \textcolor{green}{$\checkmark$} & \textcolor{green}{$\checkmark$} & \textcolor{green}{$\checkmark$} & Direct/Diagnostic \\
\bottomrule
\end{tabular}
\end{table*}

\subsection{Language-Guided Place Recognition}

Natural language supports place retrieval through descriptions of landmarks, scene attributes, and spatial relations. Text4VPR associates multi-directional street views with language descriptions \cite{shang2025text4vpr}, while UniLoc investigates bidirectional retrieval across images, text, and point clouds \cite{xia2024uniloc}. LaVPR provides large-scale language annotations and evaluates descriptions against reference images distinct from their sources \cite{idan2026lavpr}. DialogueVPR extends place retrieval to interactive, multi-turn queries \cite{song2026dialoguevpr}. Complementary localization tasks include language-to-point-cloud localization in Text2Loc \cite{xia2024text2loc}, partial text--submap correspondence \cite{feng2025partialsubmap}, and text-based satellite/map retrieval in CVG-Text \cite{ye2025cvgtext}.

These works establish diverse language-grounded geographic retrieval settings. AcrossPlace focuses on bidirectional retrieval between descriptions and street-level images under unified place-identity relevance. It combines explicit description provenance with paired source--target evaluations, geographic partitions, and controlled gallery scaling to distinguish observation transfer from candidate competition. Table~\ref{tab1} compares the documented evaluation protocols of closely related resources.

\subsection{Fine-Grained Vision--Language Understanding}

Language-guided recognition also depends on fine-grained visual grounding and geographic distinctiveness. FLAIR learns localized representations from detailed descriptions for fine-grained retrieval \cite{xiao2025flair}. Compositional grounding investigates object attributes and relations \cite{zeng2024compositional}, while SpatialVLM and SpatiaLQA address spatial understanding and reasoning \cite{chen2024spatialvlm,xie2026spatialqa}. For place recognition, a description may be correct for its source image yet contain evidence absent from a target observation or shared by other locations. AcrossPlace evaluates these distinct properties through a three-axis audit of source correctness, target visibility, and place specificity, linking language evidence to retrieval outcomes.

\subsection{Retrieval Reference Models}

Pretrained vision--language models provide general-purpose image--text representations. CLIP \cite{radford2021clip} and SigLIP2 \cite{tschannen2025siglip2} learn aligned cross-modal embeddings, while BLIP combines contrastive features with image--text matching \cite{li2022blip, ren2025minima, stoica2025contrastive}. AcrossPlace evaluates these references alongside place-supervised adaptations, parameter-efficient fine-tuning, and validation-selected reranking under fixed retrieval protocols. The resulting comparisons characterize model behavior across observation transfer, gallery competition, and language-evidence conditions.

%% file: sec/3_benchmark.tex
\section{AcrossPlace Benchmark}

\begin{figure*}[!t]
\centering
\includegraphics[width=1.0\textwidth]{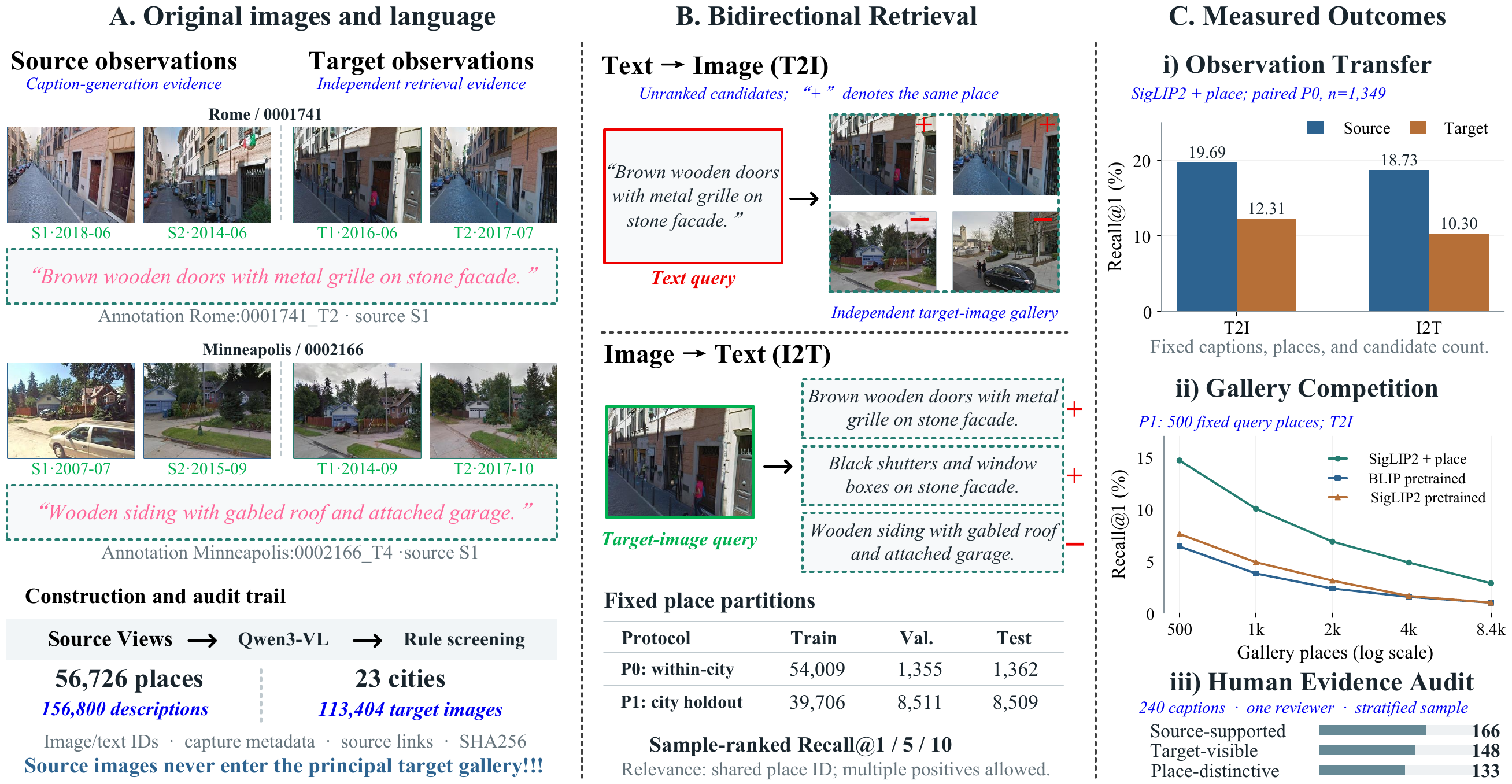}
\caption{AcrossPlace overview using original images and verbatim annotations. (a) Two places with source/target dates and traceable text. (b) Bidirectional place relevance, illustrated with unranked candidates. (c) Archived paired-P0 and fixed-query-P1 diagnostics, distinct from principal full-gallery scores; supervised results average three seeds. Human counts describe a stratified, single-reviewer audit.}\label{fig2}
\end{figure*}

\begin{figure}[t]
\centering
\includegraphics[width=1.0\columnwidth]{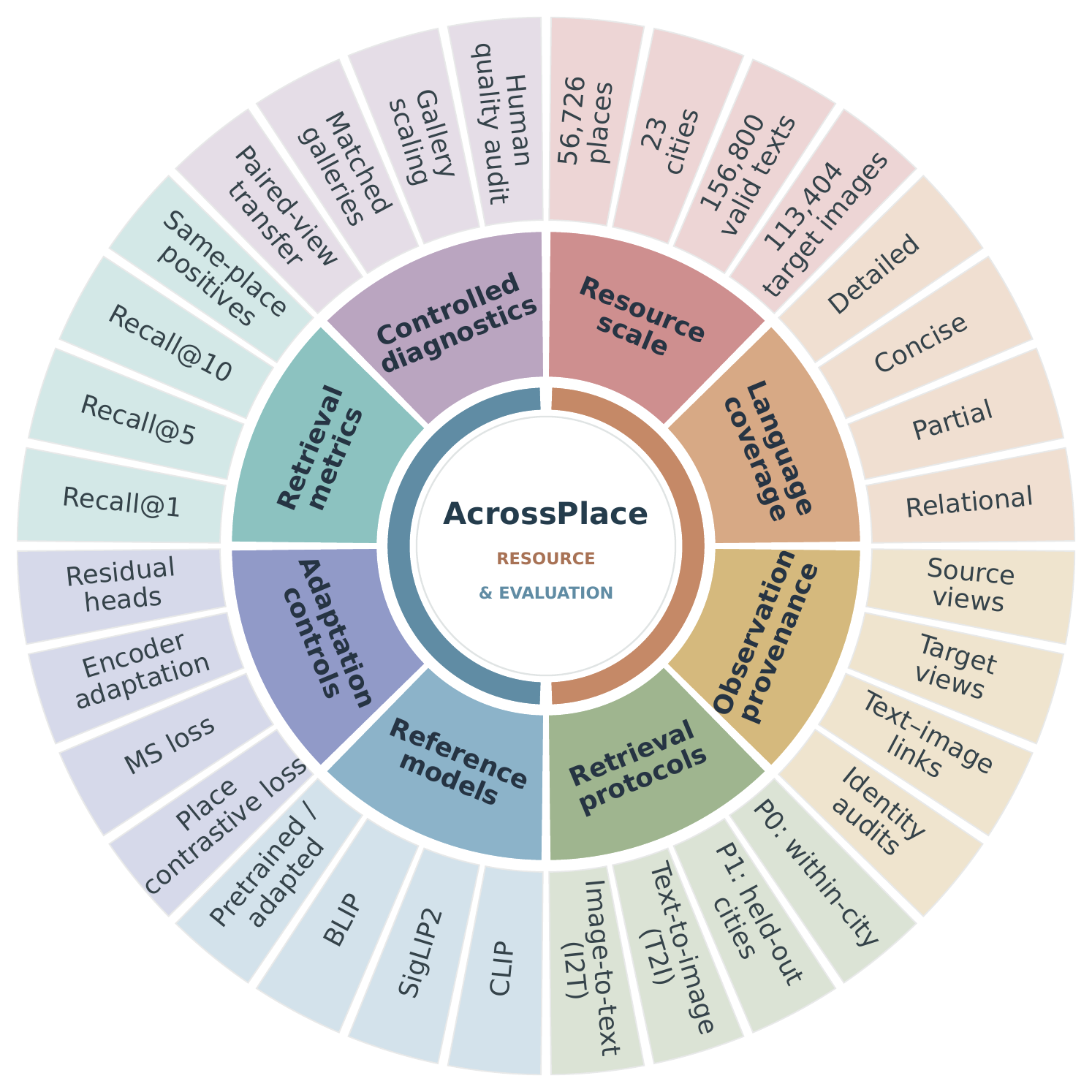}
\caption{Scope of AcrossPlace. Eight themes and 32 facets connect resources/protocols (warm/green) with models, adaptation, metrics, and diagnostics (cool/purple). Equal-angle leaves denote categories, not sample frequencies.}\label{fig3}
\end{figure}

AcrossPlace evaluates transferable place identity by linking descriptions to their source observations while separating those sources from retrieval targets under a unified place-level relevance criterion. Fixed geographic protocols and controlled diagnostics assess observation transfer, gallery competition, and language-evidence reliability (Figure~\ref{fig2}); Figure~\ref{fig3} summarizes the benchmark scope.

\subsection{Resource Construction and Observation Provenance}

\textbf{Source and target observations.}
We begin with 62,279 GSV-Cities places, each containing four images (249,116 images). Qwen3-VL-4B-Instruct generates descriptions from designated \emph{source observations}; distinct images of the same place form the \emph{target pool}. Each description retains its source-image link. Image/text identifiers, canonical paths, capture timestamps, GPS metadata, panorama identifiers, and SHA256 hashes are preserved in manifests to verify source--target separation.

\textbf{Eligible resource.}
Rule-based screening retains \textbf{56,726 places across 23 cities}, \textbf{156,800 accepted descriptions}, and \textbf{113,404 target images}. Accepted descriptions form principal text queries and galleries; rule-uncertain descriptions support quality diagnostics. Screening assesses rule compliance, while factual support is examined through human auditing. All benchmark totals refer to this eligible resource.

\textbf{Independent visual observations.}
Median source--target heading differences are 10$^\circ$ on P0 and 9$^\circ$ on P1, with capture-time gaps of 46 and 51 months, respectively. Controlled source--target pairs share neither identical image bytes nor panorama IDs. Thus, \emph{cross-view} denotes independent observations of a place, potentially with similar viewing directions, rather than necessarily large angular changes.

\subsection{Geographic Evaluation Protocols}

AcrossPlace defines two fixed, place-disjoint geographic protocols with predetermined query/gallery manifests (Table~\ref{tab2}).

\textbf{P0: Unseen places within represented cities.}
P0 tests unseen places in cities represented during training, with 54,009 training, 1,355 validation, and 1,362 test places.

\textbf{P1: Generalization to held-out cities.}
P1 separates cities across splits, assigning four to validation and four distinct cities to testing. Its training, validation, and test sets contain 39,706, 8,511, and 8,509 places. City assignments follow geographic metadata and a fixed seed.

P0 and P1 characterize within-city and held-out-city evaluation, respectively. Because training cohorts, queries, and galleries also differ, their raw score gap reflects combined protocol differences. Cardinality-matched and fixed-query controls further examine gallery effects.

\begin{table}[t]
\centering\small
\caption{Fixed geographic partitions. P0 and P1 use the same 56,726 places, 156,800 valid descriptions, and 113,404 target images, but assign places to different splits. Source observations are kept separate.}\label{tab2}
\begin{tabular}{@{}llrrr@{}}
\toprule
Protocol & Split & Places & Texts & Targets \\ \midrule
P0 & Train & 54,009 & 149,126 & 107,970 \\ P0 & Validation & 1,355 & 3,876 & 2,710 \\ P0 & Test & 1,362 & 3,798 & 2,724 \\ \midrule P1 & Train & 39,706 & 109,809 & 79,390 \\ P1 & Validation & 8,511 & 23,622 & 17,000 \\ P1 & Test & 8,509 & 23,369 & 17,014 \\
\bottomrule
\end{tabular}
\end{table}

\subsection{Bidirectional Retrieval and Place-Level Relevance}

In \textbf{text-to-image (T2I)} retrieval, descriptions query independent target images; in \textbf{image-to-text (I2T)} retrieval, target images query stored descriptions. Let $p(x)$ denote the place identity of image or description $x$. Candidate $g$ is relevant to query $q$ when $p(g)=p(q)$, regardless of their original image--caption pairing. Multiple gallery samples may therefore represent the same relevant place.

We rank \emph{individual samples}, without place deduplication, and report Recall@$K$ for $K\in\{1,5,10\}$:
\begin{equation}
R@K=\frac{1}{|Q|}\sum_{q\in Q}\mathbf{1}\!\left[\exists g\in \mathrm{TopK}(q):p_g=p_q\right].
\end{equation}
Repeated same-place items retain separate ranks. Principal galleries use the corresponding held-out split; target-image galleries exclude caption-source images. GPS and source observations support construction and auditing, not model inference; I2T retrieves rather than generates descriptions. The separate language-evidence audit (Sec.~\ref{sec:quality}) examines visual support and place specificity without altering place-ID relevance labels.

\subsection{Controlled Diagnostic Evaluations}

Three diagnostics complement principal Recall@$K$ by characterizing observation transfer (D1), gallery competition (D2), and language evidence (D3).

\textbf{D1: Observation transfer.}
We select one accepted, single-source description and one source--target pair per eligible place. For T2I, fixed descriptions query equal-sized source or independent-target galleries; for I2T, the corresponding source or target image queries a fixed caption gallery. Place cohorts and candidate cardinality remain unchanged. The diagnostic includes 1,349 P0 and 8,425 P1 places; 13 and 84 test places, respectively, fail the single-source criterion. Paired differences capture the combined effect of replacing observations, including temporal, directional, and visibility changes.

\textbf{D2: Geographic composition and gallery competition.}
We sample five 1,000-place galleries per protocol, each with one image and one description per place, to match candidate counts across P0/P1. Separately, 500 fixed query places are evaluated against nested galleries of 500, 1,000, 2,000, 4,000, and the full eligible test pool where available. Query identities, positive-place membership, and model weights remain fixed; galleries and relevance labels are predetermined. Gallery-draw and training-seed variability are reported separately.

\textbf{D3: Language-evidence reliability.}
On P1, 7,735 test places have both rule-accepted and rule-uncertain descriptions. A matched 4,291-place subset controls description type, source observation, and word-count bin while fixing query places, candidate counts, and background captions as focal description status varies. P0 has no corresponding uncertain-caption test pool. A blinded, single-reviewer audit examines 240 descriptions (120 per stratum) along three axes: \emph{source correctness}, \emph{target visibility}, and \emph{place specificity}. Together, matched retrieval and human audit distinguish screening status from visual support and geographic discriminability.

%% file: sec/4_setup.tex
\section{Evaluation Setup}

\subsection{Reference Systems and Adaptation}

We evaluate four pretrained references: CLIP ViT-B/16, CLIP ViT-B/32, BLIP contrastive features, and SigLIP2. Four place-supervised configurations include CLIP with place loss, LaVPR-inspired BLIP with Multi-Similarity (MS) loss, and SigLIP2 with MS or bidirectional place loss. Four additional SigLIP2 variants examine residual heads and modality-specific adaptation. Each supervised configuration uses three seeds (42, 43, 44) and a common three-epoch budget.

The standard setup samples 16 places per batch, with up to two target images and four descriptions per place. It employs AdamW and rank-eight LoRA on query/value projections in the final four blocks of both encoders, with identity-initialized residual heads. BLIP+MS instead uses all-linear rank-64 rank-stabilized LoRA, an image-to-text objective, and scaled SGD. These settings provide reproducible comparisons under a fixed adaptation budget.

\subsection{Checkpoint Selection and Statistical Reporting}

Checkpoints maximize the weaker-direction validation Recall@100, with weaker-direction Recall@1 breaking ties. A frozen BLIP image--text matching (ITM) model reranks coarse Top-100 candidates using validation-selected score fusion. Checkpoints, fusion weights, and evaluation settings are fixed before testing; coarse and reranked results are reported separately.

Supervised results aggregate three training seeds, while pretrained references use fixed checkpoints. Repeated gallery draws quantify candidate-sampling variability, and paired place-level bootstrap intervals assess source--target retrieval differences. A stratified, single-reviewer audit of 240 descriptions assesses source correctness, target visibility, and place specificity. These analyses support reproducible evaluation of observation transfer, gallery competition, and language-evidence reliability.

%% file: sec/5_results.tex
\section{Results}

We assess transferable place identity through P0/P1 reference retrieval and three diagnostics: observation transfer (D1), gallery competition (D2), and language-evidence reliability (D3). Training-seed variation, gallery draws, and paired-place confidence intervals quantify distinct sources of uncertainty.

\paragraph{Reference performance.}
Tables~\ref{tab3} and~\ref{tab4} compare pretrained and place-supervised references. On P0, SigLIP2 T2I/I2T R@1 improves from 4.16\%/5.29\% to 12.95\%/13.64\% with place supervision, reaching 15.24\%/17.35\% with ITM reranking. BLIP+MS achieves the highest P0 coarse R@1 (17.54\%/23.62\%), compared with 16.87\%/21.28\% after reranking. On P1, its coarse and reranked scores are 6.09\%/8.36\% and 6.23\%/8.18\%, respectively. Place supervision improves the tested references, while reranking effects vary by representation and direction.

\paragraph{Adaptation behavior.}
Within the common three-epoch budget, BLIP+MS validation T2I/I2T R@1 on P0 increases from 13.97\%/20.26\% to 16.94\%/23.75\%. These results characterize reference performance at a fixed adaptation stage, with comparisons based on retrieval metrics rather than objective-dependent loss values.
\begin{table*}[!t]
\centering
\caption{P0 test Recall@K (\%). Supervised R@1: mean $\pm$ SD across three training seeds; R@5/10: seed means. Pretrained controls use one frozen checkpoint. ITM denotes frozen BLIP~\cite{li2022blip} matching of coarse Top-100 results. MS models are common-budget adaptations, not reproductions of native training schedules.}\label{tab3}
\begin{tabular}{@{}llrrrrrr@{}}
\toprule
Reference & Scoring & \multicolumn{3}{c}{T2I} & \multicolumn{3}{c}{I2T} \\ & & R@1 & R@5 & R@10 & R@1 & R@5 & R@10 \\ \midrule
CLIP B/32 (zero-shot) & Coarse & 1.76 & 5.87 & 9.35 & 2.68 & 8.48 & 12.74 \\
CLIP B/16 (zero-shot) & Coarse & 2.13 & 5.61 & 8.77 & 3.63 & 9.58 & 13.62 \\
BLIP (zero-shot) & Coarse & 4.08 & 10.01 & 13.72 & 4.41 & 11.01 & 16.04 \\
SigLIP2 (zero-shot) & Coarse & 4.16 & 9.06 & 12.59 & 5.29 & 11.34 & 15.93 \\
SigLIP2 (zero-shot) & + ITM & 7.61 & 16.06 & 20.30 & 10.02 & 19.42 & 24.01 \\
CLIP B/16 + place loss & Coarse & 10.59 $\pm$ 0.12 & 23.98 & 33.39 & 11.17 $\pm$ 0.11 & 26.35 & 36.81 \\
CLIP B/16 + place loss & + ITM & 13.54 $\pm$ 0.29 & 27.08 & 36.40 & 15.38 $\pm$ 0.04 & 31.64 & 40.79 \\
BLIP + MS adaptation & Coarse & \bestresult{17.54 $\pm$ 0.30} & \bestresult{35.13} & \bestresult{44.89} & \bestresult{23.62 $\pm$ 0.27} & \bestresult{42.73} & \bestresult{52.80} \\
BLIP + MS adaptation & + ITM & \secondresult{16.87 $\pm$ 0.28} & \secondresult{31.60} & \secondresult{41.11} & \secondresult{21.28 $\pm$ 0.20} & \secondresult{40.44} & \secondresult{49.47} \\
SigLIP2 + MS & Coarse & 14.25 $\pm$ 0.11 & 29.31 & 38.05 & 17.35 $\pm$ 0.09 & 36.23 & \thirdresult{45.69} \\
SigLIP2 + MS & + ITM & 14.56 $\pm$ 0.03 & 29.47 & 37.34 & \thirdresult{19.20 $\pm$ 0.10} & \thirdresult{36.28} & 45.33 \\
SigLIP2 + place loss & Coarse & 12.95 $\pm$ 0.22 & 27.68 & 36.69 & 13.64 $\pm$ 0.27 & 30.87 & 41.45 \\
SigLIP2 + place loss & + ITM & \thirdresult{15.24 $\pm$ 0.30} & \thirdresult{29.73} & \thirdresult{38.12} & 17.35 $\pm$ 0.15 & 34.88 & 44.77 \\
\bottomrule
\end{tabular}
\end{table*}

\begin{table*}[!t]
\centering
\caption{P1 full-gallery test Recall@K (\%). Supervised R@1: mean $\pm$ SD over three seeds; R@5/10: means. Reranking weights are validation-selected. P1 differs from P0 in training and query cohorts as well as gallery composition and size.}\label{tab4}
\begin{tabular}{@{}llrrrrrr@{}}
\toprule
Reference & Scoring & \multicolumn{3}{c}{T2I} & \multicolumn{3}{c}{I2T} \\ & & R@1 & R@5 & R@10 & R@1 & R@5 & R@10 \\ \midrule
CLIP B/32 (zero-shot)& Coarse & 0.57 & 1.70 & 2.66 & 0.81 & 2.42 & 3.79 \\
CLIP B/16 (zero-shot)& Coarse & 0.42 & 1.51 & 2.46 & 0.92 & 2.66 & 4.08 \\
BLIP (zero-shot)& Coarse & 1.14 & 3.30 & 4.97 & 1.10 & 3.27 & 4.91 \\
SigLIP2 (zero-shot)& Coarse & 1.20 & 3.54 & 5.23 & 1.42 & 3.72 & 5.30 \\
SigLIP2 (zero-shot)& + ITM & 2.75 & 6.35 & 8.60 & 2.66 & 6.49 & 8.80 \\
CLIP + place & Coarse & 2.70 $\pm$ 0.08 & 7.80 & 11.60 & 2.82 $\pm$ 0.05 & 8.02 & 11.90 \\
CLIP + place & + ITM & 4.31 $\pm$ 0.05 & 10.34 & 14.23 & 4.70 $\pm$ 0.03 & 11.54 & 15.82 \\
BLIP + MS & Coarse & \secondresult{6.09 $\pm$ 0.12} & \bestresult{14.30} & \bestresult{19.66} & \bestresult{8.36 $\pm$ 0.10} & \bestresult{18.23} & \bestresult{24.38} \\
BLIP + MS & + ITM & \bestresult{6.23 $\pm$ 0.02} & \secondresult{13.87} & \secondresult{19.00} & \secondresult{8.18 $\pm$ 0.09} & \secondresult{17.63} & \secondresult{23.68} \\
SigLIP2 + MS & Coarse & 4.11 $\pm$ 0.04 & 10.31 & 14.84 & 5.18 $\pm$ 0.04 & 12.91 & 18.10 \\
SigLIP2 + MS & + ITM & \thirdresult{5.25 $\pm$ 0.04} & 11.92 & \thirdresult{16.47} & \thirdresult{6.40 $\pm$ 0.04} & \thirdresult{14.79} & \thirdresult{20.07} \\
SigLIP2 + place & Coarse & 3.68 $\pm$ 0.05 & 9.52 & 13.74 & 3.46 $\pm$ 0.04 & 9.97 & 14.59 \\
SigLIP2 + place & + ITM & 5.21 $\pm$ 0.05 & \thirdresult{12.03} & \thirdresult{16.47} & 5.72 $\pm$ 0.08 & 13.38 & 18.25 \\
\bottomrule
\end{tabular}
\end{table*}

\subsection{Finding I: Recognition degrades across independent observations}

\begin{figure}[tp]
\centering
\includegraphics[width=1.0\linewidth]{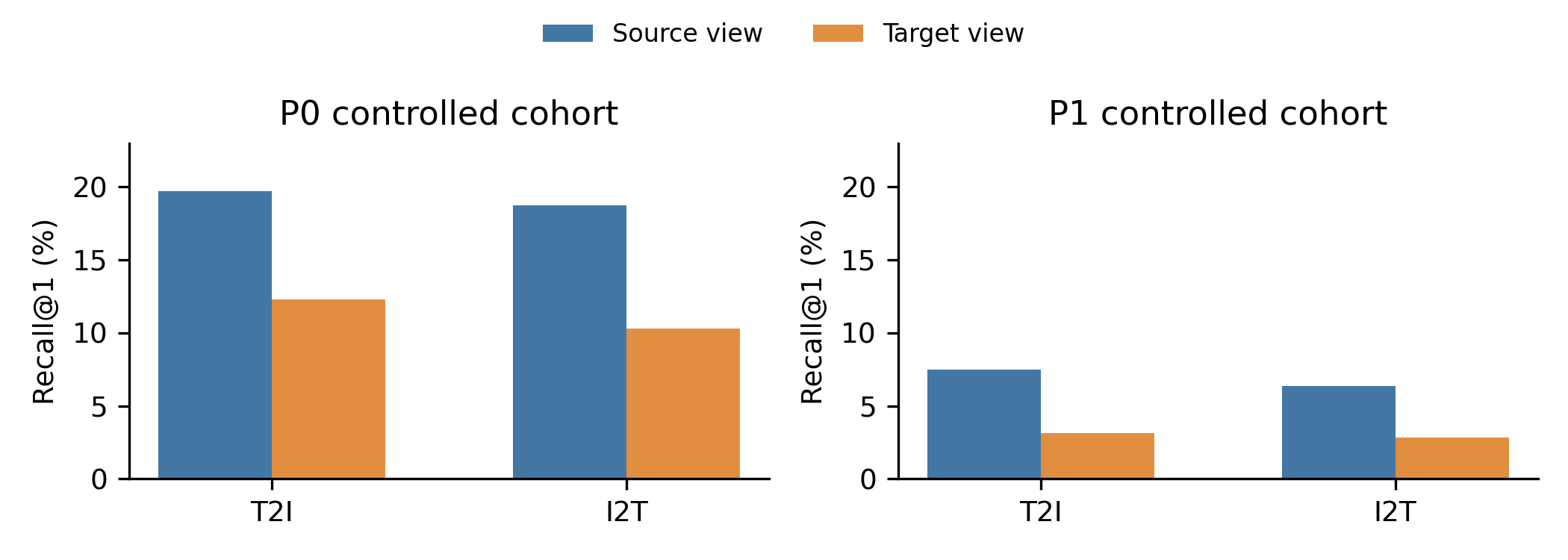}
\caption{Observation-transfer diagnostic. Identical captions and place cohorts are retrieved against equal-size source or independent-target galleries. Bars average three supervised SigLIP2 seeds; this one-sample-per-place setting differs from principal retrieval.}\label{fig4}
\end{figure}

D1 tests observation transfer by fixing place cohorts, descriptions, and gallery cardinality while replacing source images with independent target observations for T2I, or switching the image query for I2T (Figure~\ref{fig4}, Table~\ref{tab5}). Each place contributes one image and one description.

For place-supervised SigLIP2 on P0, T2I R@1 falls from 19.69\% to 12.31\% and I2T from 18.73\% to 10.30\%, with paired-bootstrap 95\% intervals of $[-9.24,-5.49]$ and $[-10.16,-6.67]$ percentage points for target-minus-source differences. On P1, T2I declines from 7.49\% to 3.11\% and I2T from 6.35\% to 2.85\%, with intervals of $[-4.92,-3.84]$ and $[-4.01,-3.01]$. Pretrained BLIP shows comparable declines, whereas the P0 pretrained SigLIP2 T2I interval $[-2.15,+0.52]$ includes zero. These results identify an \emph{observation-transfer gap}: agreement with caption-source images does not reliably predict recognition from independent evidence. The paired comparison measures the combined effects of observation changes, and its intervals reflect place resampling rather than training-seed variation.

\begin{table*}[!t]
\centering\small
\setlength{\tabcolsep}{4pt}
\caption{Source-to-target observation transfer (\%; $\Delta$ in percentage points). Same captions and equal-size galleries for 1,349 P0 and 8,425 P1 places. 95\% CIs use 2,000 paired place bootstraps after averaging seed outcomes; intervals exclude training-seed uncertainty.}\label{tab5}
\begin{tabular}{@{}llrrrrrr@{}}
\toprule
Protocol & Reference & \multicolumn{3}{c}{T2I R@1} & \multicolumn{3}{c}{I2T R@1} \\ & & Source & Target & $\Delta$ [95\% CI] & Source & Target & $\Delta$ [95\% CI] \\ \midrule
P0 & BLIP (pretrained) & \secondresult{7.78} & \thirdresult{3.48} & -4.30 [-5.78, -2.82] & \secondresult{6.89} & \thirdresult{3.63} & -3.26 [-4.52, -2.00] \\
P0 & SigLIP2 (pretrained) & \thirdresult{5.71} & \secondresult{4.89} & -0.82 [-2.15, +0.52] & \thirdresult{6.08} & \secondresult{4.60} & -1.48 [-2.82, -0.22] \\
P0 & SigLIP2 (place-supervised) & \bestresult{19.69} & \bestresult{12.31} & -7.39 [-9.24, -5.49] & \bestresult{18.73} & \bestresult{10.30} & -8.43 [-10.16, -6.67] \\
P1 & BLIP (pretrained) & \thirdresult{2.85} & \thirdresult{1.12} & -1.73 [-2.12, -1.35] & \thirdresult{2.49} & \thirdresult{1.04} & -1.45 [-1.82, -1.09] \\
P1 & SigLIP2 (pretrained) & \secondresult{2.99} & \secondresult{1.27} & -1.72 [-2.10, -1.36] & \secondresult{2.56} & \secondresult{1.13} & -1.44 [-1.76, -1.10] \\
P1 & SigLIP2 (place-supervised) & \bestresult{7.49} & \bestresult{3.11} & -4.37 [-4.92, -3.84] & \bestresult{6.35} & \bestresult{2.85} & -3.50 [-4.01, -3.01] \\
\bottomrule
\end{tabular}
\end{table*}

\subsection{Finding II: Gallery competition shapes geographic comparisons}

D2 distinguishes candidate competition from broader P0/P1 protocol differences through cardinality-matched galleries and fixed-query gallery scaling. At 1,000 places (Table~\ref{tab6}), supervised SigLIP2 achieves T2I/I2T R@1 of 13.58\%/11.69\% on P0 versus 10.61\%/9.37\% on P1. Pretrained BLIP instead scores 3.88\% on P0 versus 4.12\% on P1 for T2I, showing that the direction of the P0--P1 difference varies by reference model. Matched cardinality controls gallery size but retains differences in training cohorts, queries, and candidate semantics.

\begin{figure*}[!t]
\centering
\includegraphics[width=0.9\linewidth]{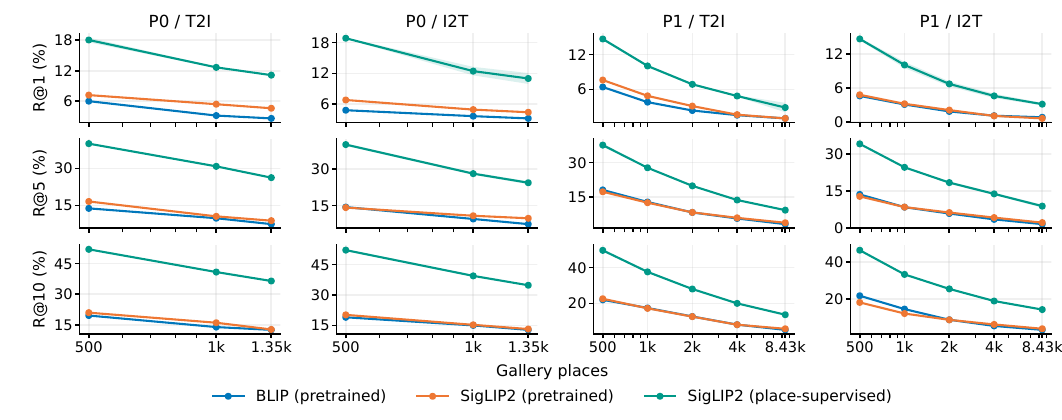}
\caption{Fixed-query gallery-scale diagnostics across protocols, directions, and ranks. Each panel uses 500 query places and five nested galleries; supervised bands show seed SD. Full-pool repeats are identical; P0 and P1 end at 1,349 and 8,425 places.}\label{fig5}
\end{figure*}

With 500 P1 query places fixed (Figure~\ref{fig5}), expanding the gallery from 500 to 8,425 places reduces supervised SigLIP2 T2I R@1 from 14.67\% to 2.87\% and I2T from 14.73\% to 3.13\%. Intermediate T2I scores are 10.03\%, 6.87\%, and 4.85\% at 1,000, 2,000, and 4,000 places. Because queries, positive identities, and model weights remain unchanged, the decline directly characterizes gallery-size sensitivity. Candidate competition is therefore essential to interpreting geographic retrieval comparisons.

\begin{table}[!t]
\centering
\centering
\caption{Matched 1,000-place gallery Recall@1 (\%), averaged over five gallery draws and, when applicable, three training seeds. Other P0/P1 differences remain.}\label{tab6}
\begin{tabular}{lrrrr}
\toprule
Reference & \multicolumn{2}{c}{P0} & \multicolumn{2}{c}{P1} \\ & T2I & I2T & T2I & I2T \\ \midrule
Pretrained BLIP & \thirdresult{3.88} & \thirdresult{4.00} & \secondresult{4.12} & \secondresult{3.40} \\
Pretrained SigLIP2 & \secondresult{5.36} & \secondresult{5.08} & \thirdresult{3.64} & \thirdresult{3.14} \\
Supervised SigLIP2 & \bestresult{13.58} & \bestresult{11.69} & \bestresult{10.61} & \bestresult{9.37} \\
\bottomrule
\end{tabular}
\end{table}

\begin{table}[t]
\centering\small
\caption{Blinded audit of 240 captions (120 per automatic stratum) by one reviewer. Unknown is distinct from contradiction or nonvisibility; counts are sample-specific.}\label{tab7}
\begin{tabular}{@{}llrrr@{}}
\toprule
Axis & Judgment & All & Valid & Uncertain \\ \midrule
Source correctness & Supported & 166 & 81 & 85 \\ & Partial & 8 & 2 & 6 \\ & Contradicted & 23 & 14 & 9 \\ & Unknown & 43 & 23 & 20 \\ \midrule Target visibility & Visible & 148 & 75 & 73 \\ & Partial & 9 & 2 & 7 \\ & Not visible & 37 & 17 & 20 \\ & Unknown & 46 & 26 & 20 \\ \midrule Specificity & Distinctive & 133 & 63 & 70 \\ & Generic & 42 & 23 & 19 \\ & Unknown & 65 & 34 & 31 \\
\bottomrule
\end{tabular}
\end{table}

\subsection{Finding III: Automated screening and language evidence capture distinct properties}\label{sec:quality}

D3 distinguishes place-ID retrieval from the visual support and specificity of language descriptions. A blinded audit covers 240 descriptions from 237 P1 places, equally divided between rule-valid and rule-uncertain strata (Table~\ref{tab7}). Of these, 166 are fully source-supported, 148 have sufficient target evidence, and 133 are place-distinctive. Full source support occurs in 81/120 valid and 85/120 uncertain descriptions, indicating that rule status does not consistently rank factual support in this sample.

In a matched 4,291-place P1 cohort, valid descriptions outperform uncertain ones in supervised SigLIP2 R@1: 3.16\% versus 2.19\% for T2I and 3.25\% versus 1.78\% for controlled I2T. Audited T2I R@10 is 17.79\% for target-visible versus 0.00\% for not-visible descriptions, and 17.07\% for source-supported versus 4.35\% for contradicted descriptions. These descriptive subgroup comparisons reflect overlapping samples and a single reviewer; partial and unknown judgments remain distinct from negative evidence.

Audited T2I searches 17,014 target images, whereas controlled I2T replaces one caption per focal place in an 8,509-caption gallery. These units differ from principal I2T retrieval. Errors involving similar facades, vegetation, and entrances illustrate semantic aliasing across distinct places. Overall, D3 separates automated screening, factual support, target visibility, and place specificity as complementary properties of language-guided recognition.

%% file: sec/6_discussion.tex
\section{Conclusion}

AcrossPlace investigates whether language--vision models recognize geographic places beyond the visual observations used to generate their descriptions. By integrating provenance-aware bidirectional retrieval with controlled diagnostics, it systematically examines three factors governing place-recognition performance: observation transfer, gallery competition, and language-evidence reliability. Paired evaluations reveal substantial retrieval degradation when caption-source images are replaced by independent observations, highlighting the distinction between image--text alignment and transferable place identity. Fixed-query gallery scaling demonstrates the influence of candidate competition, while geographic comparisons underscore the importance of controlling gallery composition. Human auditing further distinguishes automated caption acceptance from factual support, target visibility, and place specificity. Together, these findings establish AcrossPlace as a reproducible framework for interpreting cross-modal place-retrieval performance under well-defined evaluation conditions. 